\documentclass{article}
\usepackage{spconf,amsmath,graphicx,hyperref}
\usepackage{booktabs}
\usepackage[table]{xcolor}
\usepackage{bm}
\newcommand{\tabnum}[1]{{\footnotesize $#1$}}
\newcommand{\tabbold}[1]{{\footnotesize $\bm{#1}$}}
\newcommand{\graynum}[1]{{\footnotesize\color{gray!80}$#1$}}

\title{VPRune: Efficient Training-free Pre-LLM Visual Token Pruning}
\name{Guangchuan Lv$^{1}$, Dianxing Shi$^{2}$\sthanks{Corresponding author: dianxingshi10@gmail.com}, Dingjie Fu$^{3}$}
\address{$^{1}$ Northeastern University, $^{2}$ Beihang University, $^{3}$ Huazhong University of Science and Technology}
\begin{document}
%
\maketitle
\begin{abstract}
Visual token pruning is a promising approach to reducing the inference cost of large vision-language models (LVLMs), yet aggressive token reduction often causes substantial performance degradation. We identify three key factors behind this degradation: text-guided selection bias, information loss from discarded tokens,
and positional distortion caused by sequence compaction. Based on these observations, we propose \textbf{VPRune}, a training-free pre-LLM pruning framework consisting of visual-only diversity selection, similarity-guided token recycling, and position-preserving restoration. Experiments on FastVLM-1.5B across multiple vision-language benchmarks demonstrate that VPRune achieves a favorable
accuracy--compression trade-off, with particularly pronounced
advantages under aggressive compression. Furthermore,  evaluations on edge-device show that VPRune effectively reduces end-to-end inference latency while maintaining superior task performance, demonstrating its practicality for resource-constrained LVLM deployment.
\end{abstract}
\begin{keywords}
Vision-language models, visual token pruning, token reduction, efficient inference, edge AI
\end{keywords}
\section{Introduction}
\label{sec:intro}

Large vision-language models (LVLMs) achieve strong multimodal understanding capability by jointly processing visual and textual tokens~\cite{blip2,llava,fastvlm}. However, visual representations often contain substantial redundancy, while long visual-token sequences increase inference latency and hinder deployment on resource-constrained edge devices. To this end, visual token pruning alleviates this cost by removing redundant visual tokens aggressively ~\cite{visionzip,cdpruner}, but such reduction often causes severe performance degradation, particularly when pruning is performed before the LLM.

\begin{figure}[t]
    \centering
    \includegraphics[width=\columnwidth]{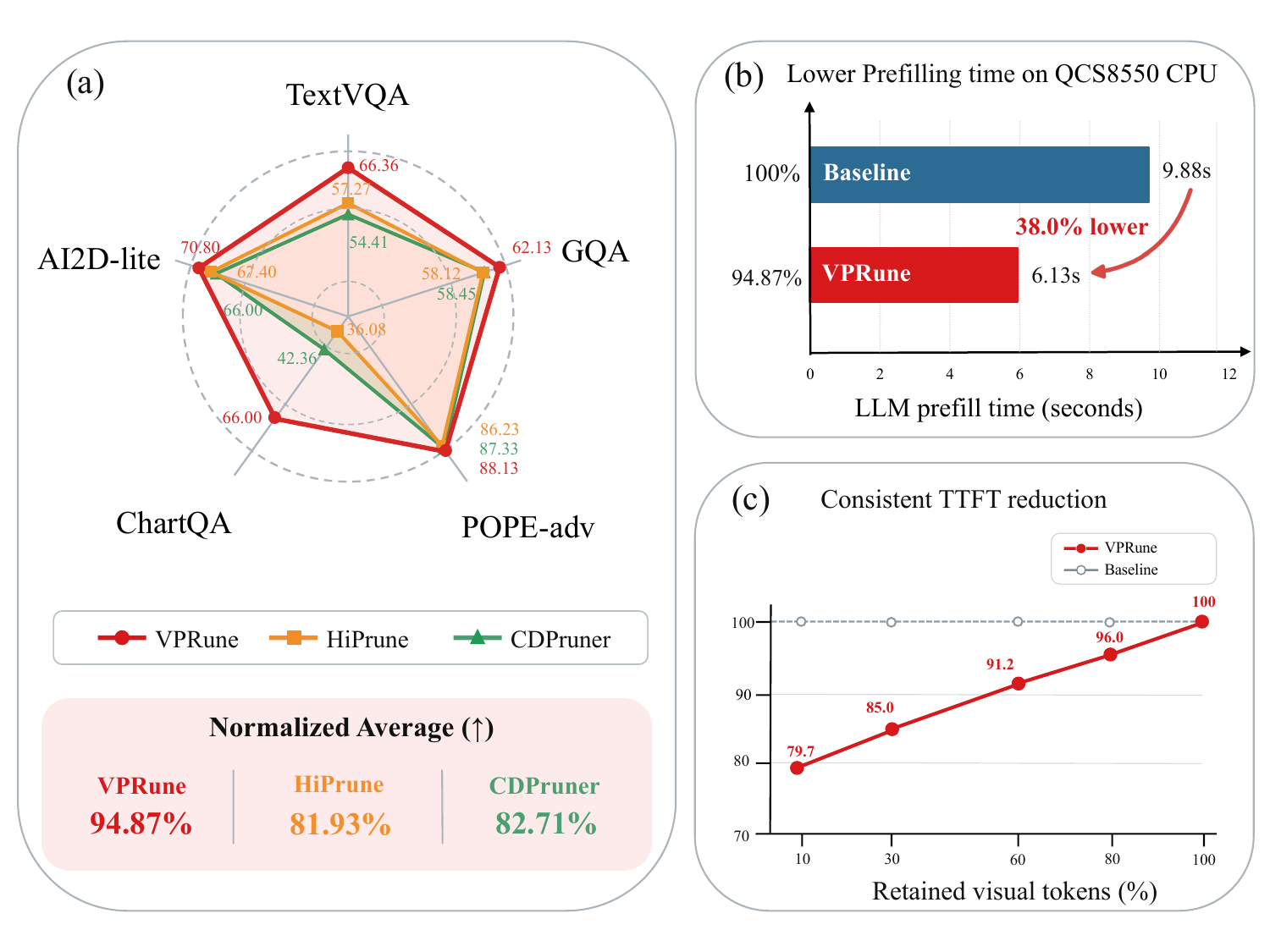}
\caption{
Performance and edge efficiency of VPRune. At 30\% token retention, VPRune achieves the highest normalized average performance of 94.87\% in (a), while reducing LLM prefill latency by 38.0\% on the QCS8550 CPU in (b); (c) further shows consistent end-to-end TTFT reductions across different visual-token budgets.
}
    \label{fig:highlight}
\end{figure}

Existing visual-token reduction methods generally improve efficiency by
removing redundant tokens before the LLM or within early decoder
layers~\cite{visionzip,hiprune,vscan,chen2024fastv}. To identify informative
tokens, they typically rely on visual redundancy, attention responses, or
cross-modal relevance~\cite{cdpruner,flashvlm}. However, many existing pruning methods are developed around token-rich
LVLM settings, while increasingly efficient edge-oriented models are
designed with compact multimodal representations for practical
deployment~\cite{fastvlm,bluelmv,tokenpacker}. In this regime, applying the same retention ratio yields a much tighter
absolute token budget, making further pruning more sensitive to information
loss and representation distortion.


To understand these degradations, we analyze pre-LLM pruning under
tight token budgets and identify three empirical observations:
\textbf{i)} text-conditioned relevance can over-concentrate token
selection and reduce visual coverage,
\textbf{ii)} discarded tokens are often redundant rather than
uninformative and may still contain complementary visual evidence, and
\textbf{iii)} sequence compaction reassigns retained tokens to
consecutive position indices, distorting the spatial relationships
seen by the pretrained LLM. Motivated by these observations, we
propose \textbf{VPRune}, a training-free pre-LLM framework with three
corresponding components. Specifically, we first introduce \textit{Visual-only Diversity Selection}
(VDS) to removes text conditioning and preserves visual diversity. We then employ \textit{Similarity-guided Token Recycling} (STR) to transfer
complementary information from discarded tokens to similar retained
tokens. Moreover, we propose \textit{Position-Preserving Restoration} (PPR) to preserve
their original raster order and position indices. Together, these
components preserve visual evidence and positional structure under
high compression without modifying the underlying LVLM.

As highlighted in Fig.~\ref{fig:highlight}, VPRune provides a favorable
accuracy--efficiency trade-off under aggressive compression.
At 30\% token retention, it achieves a normalized average performance
of 94.87\%, substantially outperforming representative 
methods. At the same retention ratio, VPRune reduces LLM prefill
latency from 9.88\,s to 6.13\,s (38.0\%) on a Qualcomm QCS8550
platform, while also providing consistent TTFT reductions across
different token budgets.

Our contributions are: \textbf{i)} We analyze three key causes of performance degradation under aggressive visual-token pruning. \textbf{ii)} We propose VPRune, a training-free pre-LLM framework integrating VDS, STR, and PPR to advance 
low-budget token reduction performance. \textbf{iii)} Extensive experiments demonstrate strong accuracy under aggressive compression together with practical prefill and end-to-end latency reductions on a CPU-only edge platform.

\section{Related Work}
\label{sec:related}

Modern LVLMs typically employ a vision encoder, multimodal projector, and LLM to jointly process visual and textual representations~\cite{blip2,llava}. Efficient architectures such as FastVLM~\cite{fastvlm} reduce visual-processing overhead, while visual token reduction provides a complementary way to remove redundant representations. Existing methods prune tokens at different stages: HiPrune~\cite{hiprune} and VisionZip~\cite{visionzip} exploit visual structure or attention patterns, CDPruner~\cite{cdpruner} combines diversity-aware DPP selection with text-conditioned relevance, FlashVLM~\cite{flashvlm} performs efficient pre-LLM filtering, VScan~\cite{vscan} incorporates query-aware token reduction, and FastV~\cite{chen2024fastv} prunes tokens after early LLM processing. Despite their effectiveness, preserving visual content, positional consistency, and discarded information remains challenging under aggressive compression. VPRune addresses these issues entirely before LLM computation.

\begin{table*}[htbp]
\centering
\caption{
Performance comparison on FastVLM-1.5B under different visual-token reduction ratios.
Baseline denotes the full-token model; ``Reduce (\%)'' is the percentage of visual tokens removed relative to it.
``Avg.'' is the mean of the five benchmark scores below normalized by their respective full-token baselines (Baseline $=100\%$).
Best and second-best results among reduction methods in each column are highlighted in bold and underlined, respectively.
}
\vspace{0.3em}
\label{tab:main_results}
\setlength{\tabcolsep}{3pt}
\renewcommand{\arraystretch}{1.2}
\resizebox{0.98\textwidth}{!}{
\begin{tabular}{
l
ccc
ccc
ccc
ccc
ccc
ccc
}
\toprule

\textbf{Method}
& \multicolumn{3}{c}{\textbf{TextVQA}}
& \multicolumn{3}{c}{\textbf{GQA}}
& \multicolumn{3}{c}{\textbf{POPE-adv}}
& \multicolumn{3}{c}{\textbf{AI2D-lite}}
& \multicolumn{3}{c}{\textbf{ChartQA}}
& \multicolumn{3}{c}{\textbf{Avg.}} \\

\cmidrule(l{3pt}r{3pt}){2-4}
\cmidrule(l{3pt}r{3pt}){5-7}
\cmidrule(l{3pt}r{3pt}){8-10}
\cmidrule(l{3pt}r{3pt}){11-13}
\cmidrule(l{3pt}r{3pt}){14-16}
\cmidrule(l{3pt}r{3pt}){17-19}

\textbf{Reduce (\%)}
& 20 & 40 & 70
& 20 & 40 & 70
& 20 & 40 & 70
& 20 & 40 & 70
& 20 & 40 & 70
& 20 & 40 & 70 \\

\midrule

{\color{gray!80}Baseline}
& \multicolumn{3}{c}{\graynum{70.58}}
& \multicolumn{3}{c}{\graynum{63.53}}
& \multicolumn{3}{c}{\graynum{87.97}}
& \multicolumn{3}{c}{\graynum{73.00}}
& \multicolumn{3}{c}{\graynum{77.32}}
& \multicolumn{3}{c}{\graynum{100\,(\%)}} \\

\midrule

CDPruner
& \tabnum{67.08} & \tabnum{64.17} & \tabnum{54.41}
& \tabnum{61.85} & \tabnum{60.91} & \tabnum{58.45}
& \tabnum{87.77} & \tabnum{87.50} & \underline{\tabnum{87.33}}
& \tabnum{71.00} & \tabnum{68.80} & \tabnum{66.00}
& \tabnum{63.88} & \tabnum{57.28} & \tabnum{42.36}
& \tabnum{94.41} & \tabnum{90.92} & \tabnum{82.71} \\

VScan
& \tabnum{67.62} & \tabnum{55.90} & \tabnum{32.06}
& \tabnum{63.21} & \tabnum{60.64} & \tabnum{54.71}
& \tabnum{87.53} & \tabnum{84.30} & \tabnum{77.57}
& \underline{\tabnum{72.20}} & \tabnum{69.80} & \tabnum{66.40}
& \tabbold{76.96} & \tabnum{57.04} & \tabnum{39.84}
& \tabnum{98.65} & \tabnum{87.97} & \tabnum{72.44} \\

FastV
& \tabnum{69.19} & \tabnum{66.67} & \tabnum{54.05}
& \tabbold{63.49} & \underline{\tabnum{62.97}} & \underline{\tabnum{58.69}}
& \underline{\tabnum{88.10}} & \underline{\tabnum{87.93}} & \tabnum{83.23}
& \tabnum{71.20} & \tabnum{70.00} & \tabnum{66.40}
& \tabnum{75.64} & \underline{\tabnum{73.64}} & \underline{\tabnum{54.36}}
& \underline{\tabnum{98.70}} & \underline{\tabnum{96.93}} & \underline{\tabnum{84.97}} \\

HiPrune
& \tabnum{69.17} & \underline{\tabnum{67.73}} & \tabnum{57.27}
& \tabnum{62.33} & \tabnum{61.15} & \tabnum{58.12}
& \tabbold{88.23} & \tabbold{88.13} & \tabnum{86.23}
& \tabnum{71.40} & \underline{\tabnum{70.40}} & \underline{\tabnum{67.40}}
& \tabnum{67.40} & \tabnum{61.08} & \tabnum{36.08}
& \tabnum{96.28} & \tabnum{93.57} & \tabnum{81.93} \\

VisionZip
& \tabnum{56.82} & \tabnum{55.04} & \tabnum{39.78}
& \tabnum{59.29} & \tabnum{59.17} & \tabnum{57.64}
& \tabnum{87.77} & \tabnum{87.63} & \tabnum{86.43}
& \tabnum{65.60} & \tabnum{66.40} & \tabnum{65.20}
& \tabnum{35.12} & \tabnum{35.28} & \tabnum{29.12}
& \tabnum{81.78} & \tabnum{81.46} & \tabnum{74.46} \\

FlashVLM
& \underline{\tabnum{69.58}} & \tabnum{67.54} & \underline{\tabnum{57.87}}
& \tabnum{62.39} & \tabnum{60.85} & \tabnum{57.17}
& \tabnum{88.07} & \tabnum{87.77} & \tabnum{85.93}
& \tabnum{71.20} & \tabnum{70.00} & \tabnum{66.40}
& \tabnum{68.12} & \tabnum{62.08} & \tabnum{35.20}
& \tabnum{96.51} & \tabnum{93.49} & \tabnum{81.23} \\

\midrule

\rowcolor{blue!3}
VPRune (Ours)
& \tabbold{70.57} & \tabbold{69.64} & \tabbold{66.36}
& \underline{\tabnum{63.29}} & \tabbold{63.07} & \tabbold{62.13}
& \tabnum{88.00} & \underline{\tabnum{87.93}} & \tabbold{88.13}
& \tabbold{72.40} & \tabbold{70.80} & \tabbold{70.80}
& \underline{\tabnum{76.20}} & \tabbold{74.08} & \tabbold{66.00}
& \tabbold{99.47} & \tabbold{98.14} & \tabbold{94.87} \\

\bottomrule
\end{tabular}}
\end{table*}

\section{Methodology}
\label{sec:method}

\begin{figure}[t]
    \centering
    \includegraphics[width=\columnwidth]{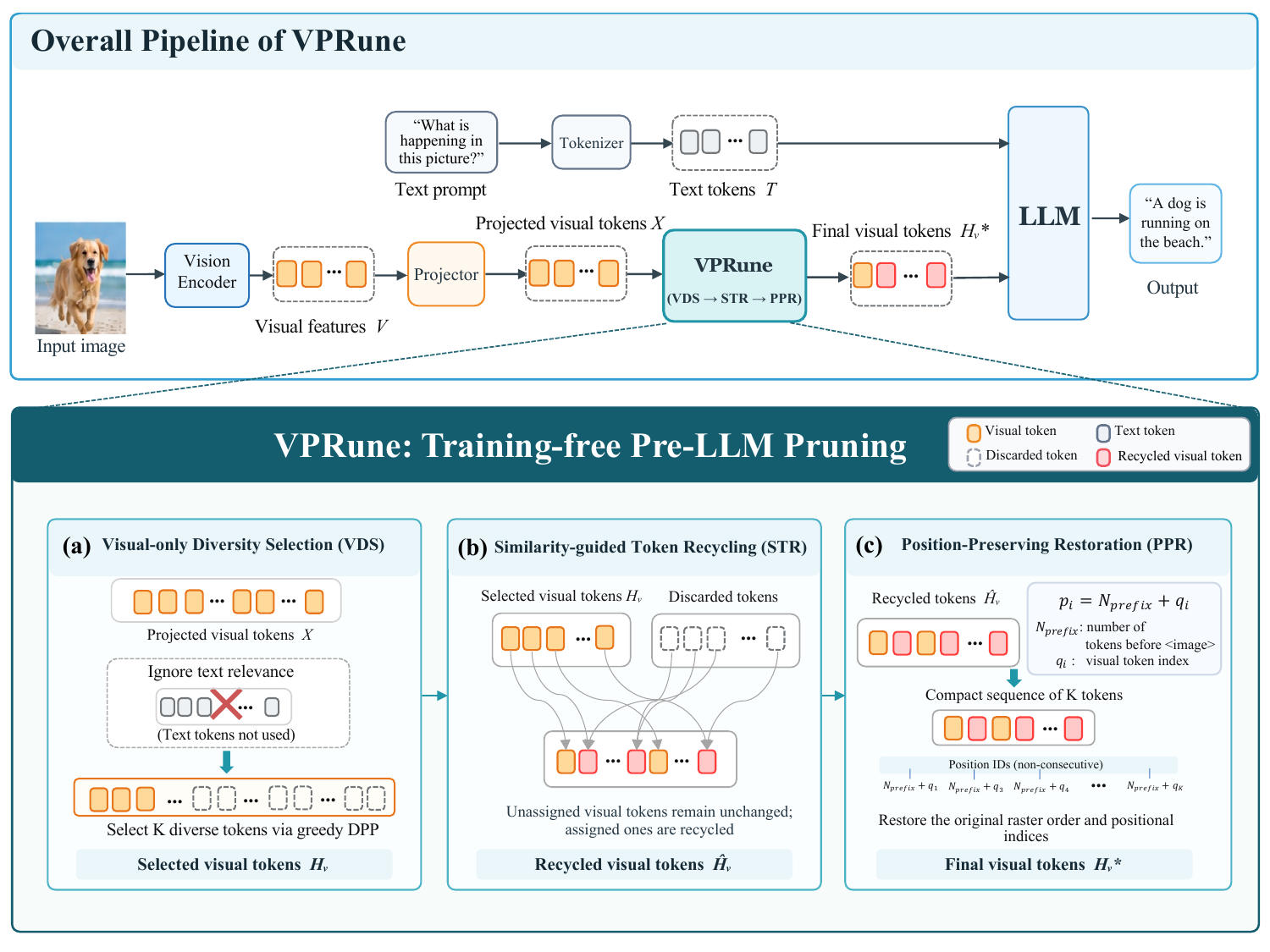}
    \caption{
    Overview of VPRune. VDS selects diverse visual tokens, STR recycles complementary information from discarded tokens, and PPR restores their original positional structure before LLM inference.
    }
    \label{fig:framework}
\end{figure}

\subsection{Overview}
\label{sec:overview}

As illustrated in Fig.~\ref{fig:framework}, VPRune is a training-free pre-LLM visual token pruning framework designed for aggressive token budgets. Given the projected visual tokens $X=\{x_i\}_{i=1}^{N}$, where $N$ denotes the number of visual tokens. VPRune consists of three stages. First, \emph{Visual-only Diversity Selection} (VDS) removes the potentially misleading text-relevance bias and selects a compact yet diverse visual subset. Second, \emph{Similarity-guided Token Recycling} (STR) recovers complementary information from discarded tokens by aggregating them into their most similar retained tokens. Finally, \emph{Position-Preserving Restoration} (PPR) restores the original raster order and positional indices of the retained tokens before they are fed into the LLM.

For a token retention ratio $b$, we retain $K=\operatorname{round}(bN)$ visual tokens. In our setting, $N=256$, and pruning is performed after the multimodal projector and before the LLM. Thus, the LLM operates on the reduced visual sequence.

\subsection{Visual-only Diversity Selection}
\label{sec:vds}

Let $L$ denote the cosine-similarity-based visual diversity kernel and $D_r=\operatorname{diag}(\hat r_1,\ldots,\hat r_N)$ is the diagonal matrix of normalized text-relevance scores. The conditioned kernel is therefore $\widetilde L=D_rLD_r$. For a retained subset $S$ of size $K$, the corresponding DPP objective \cite{kulesza2012dpp} can be decomposed as
\begin{equation}
\log\det(\widetilde L_S)
=
\log\det(L_S)
+
2\sum_{i\in S}\log\hat r_i .
\label{eq:vds_decompose}
\end{equation}
This decomposition reveals that text relevance contributes an explicit additive bias to the diversity objective. Under aggressive compression, visually complementary tokens may therefore be strongly suppressed  because they receive low text-relevance scores. VDS removes this conditioning and performs selection solely according to visual diversity. The retained subset is obtained by
\begin{equation}
S^*
=
\arg\max_{|S|=K}
\log\det(L_S).
\label{eq:vds}
\end{equation}
We solve this objective with greedy MAP inference~\cite{chen2018fastdpp}.

\subsection{Similarity-guided Token Recycling}
\label{sec:str}

Directly removing unselected tokens may discard complementary visual information, especially under tight token budgets. STR therefore recycles discarded features into their most similar retained tokens. For each discarded token $i\in D$, we assign it to $j^*(i)=\arg\max_{j\in S}\cos(x_i,x_j)$, reusing the similarities already computed in the DPP kernel. Let $G_j=\{i\in D\mid j^*(i)=j\}$ denote the discarded tokens assigned to retained token $j$. Their contribution is weighted according to
\begin{equation}
w_{ij}
=
\frac{\exp(L_{ij}/\tau)}
{\sum_{i'\in G_j}\exp(L_{i'j}/\tau)} .
\label{eq:str_weight}
\end{equation}
Rather than introducing additional tokens, STR directly updates the retained representation as
\begin{equation}
\widetilde{x}_j
=
(1-\gamma)x_j
+
\gamma\sum_{i\in G_j}w_{ij}x_i ,
\label{eq:str}
\end{equation}
where $\gamma$ controls the amount of recycled information. 
If no discarded token is assigned to $j$, i.e., $G_j=\emptyset$, 
we simply keep $\widetilde{x}_j=x_j$. 
Thus, STR recovers complementary information without increasing 
the number of tokens passed to the LLM.

\subsection{Position-Preserving Restoration}
\label{sec:ppr}

Conventional token pruning typically compacts the retained visual sequence and reassigns consecutive position indices, altering the positional relationships encoded by RoPE-based LLMs~\cite{su2021roformer}. Let the retained tokens correspond to original raster indices $S=\{q_i\}_{i=1}^{K}$ with $q_1<\cdots<q_K$. PPR first restores their original sequence order and assigns each retained token the position
\begin{equation}
p_i^{\mathrm{vis}}
=
N_{\mathrm{prefix}}+q_i ,
\label{eq:ppr}
\end{equation}
instead of the compact position $N_{\mathrm{prefix}}+i$. Following GAP~\cite{gap}, text tokens after the image start from
$N_{\mathrm{prefix}}+N$, preserving the original visual-token span.
Importantly, the retained embeddings remain a compact $K$-token
sequence; gaps exist only in the assigned position indices rather
than as empty token slots. This preserves the original raster order
and relative positional distances without reconstructing a
length-$N$ visual sequence.

\section{Experiments}
\label{sec:experiments}

\subsection{Experimental Setup}
\label{sec:setup}

\textbf{Model and Baselines.}
We evaluate VPRune on the frozen FastVLM-1.5B model ~\cite{fastvlm} and compare it with six representative visual-token reduction methods: CDPruner~\cite{cdpruner}, VScan~\cite{vscan}, FastV~\cite{chen2024fastv}, HiPrune~\cite{hiprune}, VisionZip~\cite{visionzip}, and FlashVLM~\cite{flashvlm}.

\noindent{\textbf{Benchmarks.}}
We evaluate five complementary benchmarks: GQA~\cite{hudson2019gqa} for compositional reasoning, TextVQA~\cite{singh2019textvqa} for text-rich understanding, POPE-adv~\cite{li2023pope} for object hallucination, a 500-sample subset of AI2D~\cite{kembhavi2016ai2d} (AI2D-lite) for diagram understanding, and ChartQA~\cite{masry2022chartqa} for chart reasoning.

\noindent{\textbf{Implementation Details.}}
FastVLM-1.5B uses FastViT-HD~\cite{fastvit} and a Qwen2 LLM.
Inputs are padded to $1024\times1024$, yielding 256 visual tokens;
VPRune prunes after the projector and before the LLM.
We report accuracy at $b\in\{0.3,0.6,0.8\}$ and additionally use
$b=0.1$ for efficiency. Baselines are matched to the target ratios.
All runs use greedy decoding (batch size 1) without fine-tuning;
VPRune sets $\gamma=0.3$ and $\tau=0.1$.

\noindent{\textbf{Edge Platform.}}
Latency is measured on a Qualcomm QCS8550-based development platform using CPU-only inference, with GPU and NPU acceleration disabled. The QCS8550 integrates a 64-bit Kryo CPU with a maximum clock frequency of 3.2\,GHz, providing a representative resource-constrained platform for evaluating practical edge inference.


\subsection{Main Results}
\label{sec:main_results}

As shown in Table~\ref{tab:main_results}, VPRune achieves the highest
normalized average performance at all reported reduction ratios,
retaining 99.47\%, 98.14\%, and 94.87\% of full-token performance
at 20\%, 40\%, and 70\% reduction, respectively. More importantly,
its margin over the second-best method grows from 0.77 and 1.21
percentage points at 20\% and 40\% reduction to 9.90 points over
FastV at 70\% reduction. At this setting, VPRune ranks first on all
five benchmarks. The largest gains occur on TextVQA and ChartQA,
where it exceeds the corresponding second-best results by 8.49 and
11.64 raw points. These tasks rely heavily on fine-grained textual
or structured visual evidence and are therefore more sensitive to
information removed by aggressive pruning. This pattern is consistent with VPRune's design: VDS and STR preserve
distributed and complementary visual evidence, while PPR maintains
spatial relationships that are particularly important for structured
tasks such as ChartQA. The smaller gap on POPE-adv,
whose scores remain near the full-token baseline for several methods,
further suggests that the main benefit appears on content-dense
reasoning tasks rather than from a uniform score shift.

\subsection{Ablation Study}
\label{sec:ablation}

\begin{figure}[t]
    \centering

    \begin{minipage}[t]{0.48\columnwidth}
        \centering
        \includegraphics[width=\linewidth]{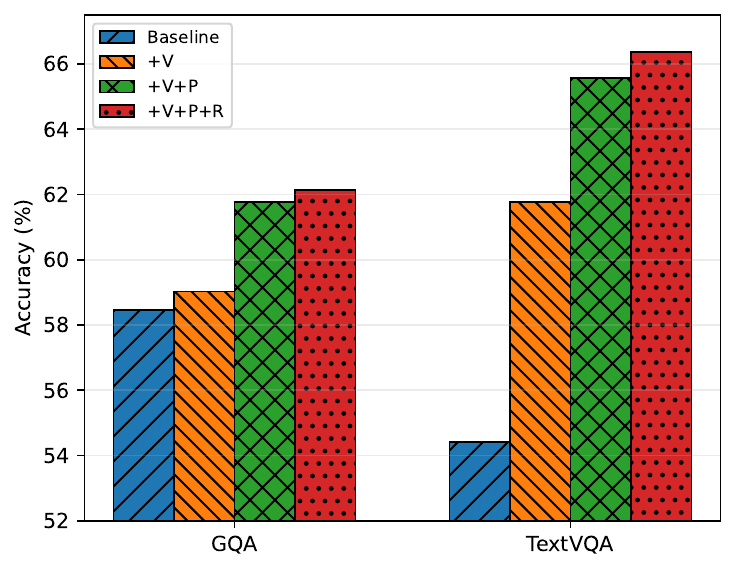}\\[-0.5mm]
        (a) Component ablation
    \end{minipage}
    \hfill
    \begin{minipage}[t]{0.48\columnwidth}
        \centering
        \includegraphics[width=\linewidth]{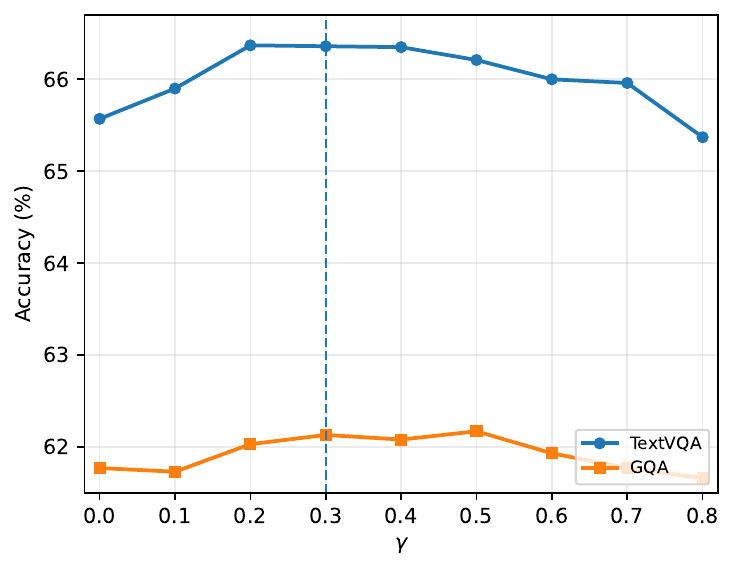}\\[-0.5mm]
        (b) Recycling strength $\gamma$
    \end{minipage}

\caption{
Ablation studies under 30\% token retention. 
(a) Cumulative ablation of V, P, and R improves performance, where V, P, and R denote visual-only diversity selection, position-preserving restoration, and similarity-guided token recycling, respectively. 
(b) $\gamma=0.3$ gives the best overall trade-off on TextVQA and GQA.
}
    \label{fig:ablation}
\end{figure}

We conduct ablations at a token retention ratio of $b=0.3$.
As shown in Fig.~\ref{fig:ablation}, removing text-guided relevance
yields a substantial gain on TextVQA and a modest improvement on GQA.
Position-preserving restoration further improves both benchmarks,
while similarity-guided token recycling provides consistent
additional gains by recovering complementary information from
discarded tokens.

We further study the effect of the recycling strength $\gamma$ on
TextVQA and GQA at $b=0.3$. Performance remains relatively stable
over moderate values of $\gamma$ but degrades when excessive
discarded information is introduced. We therefore set $\gamma=0.3$,
which provides the best overall trade-off across the two benchmarks.

\subsection{Efficiency Analysis}
\label{sec:efficiency}

\begin{figure}[t]
    \centering

    \begin{minipage}[t]{0.48\columnwidth}
        \centering
        \includegraphics[width=\linewidth]{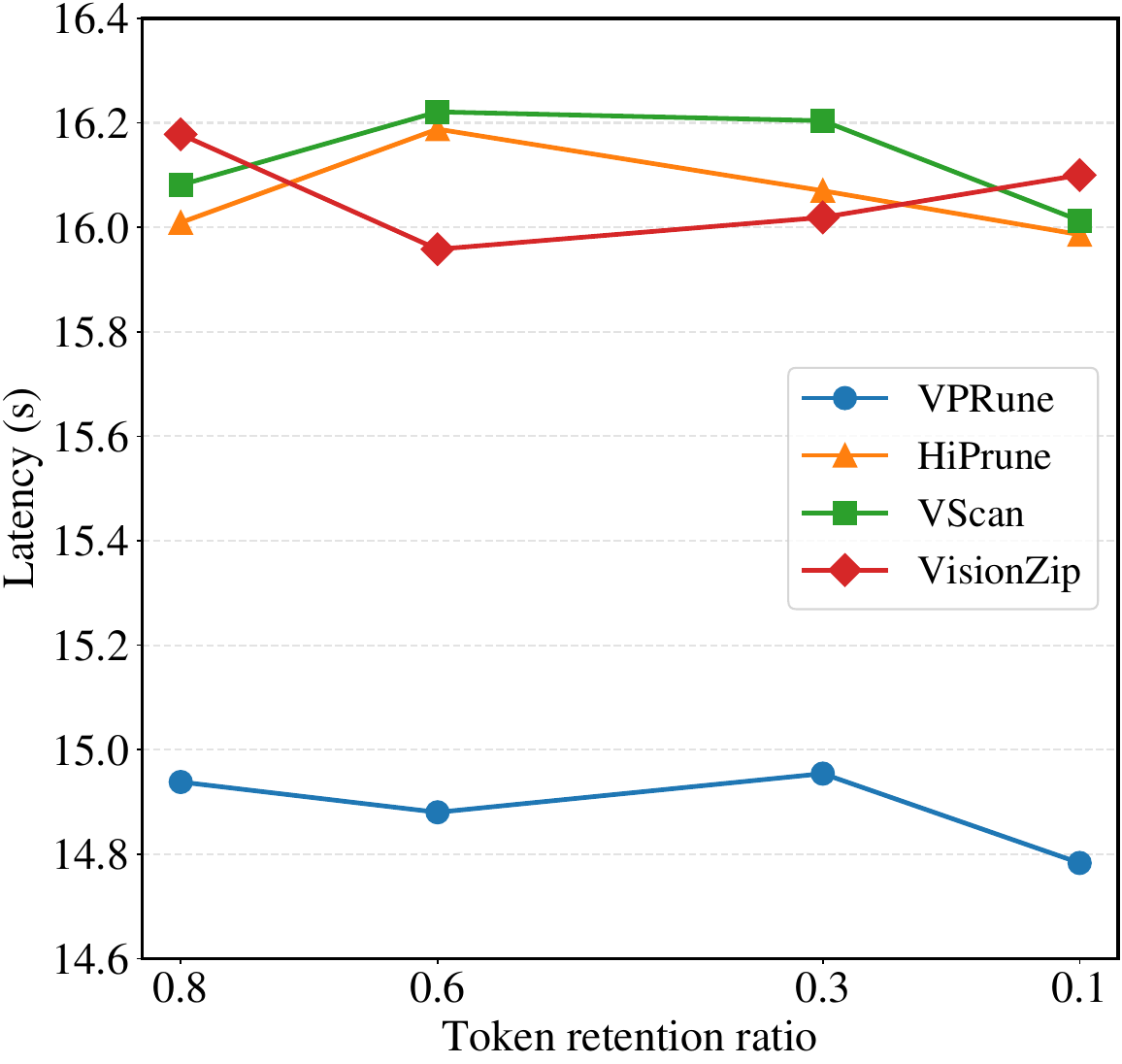}\\[-0.5mm]
        (a) Vision encoder
    \end{minipage}
    \hfill
    \begin{minipage}[t]{0.48\columnwidth}
        \centering
        \includegraphics[width=\linewidth]{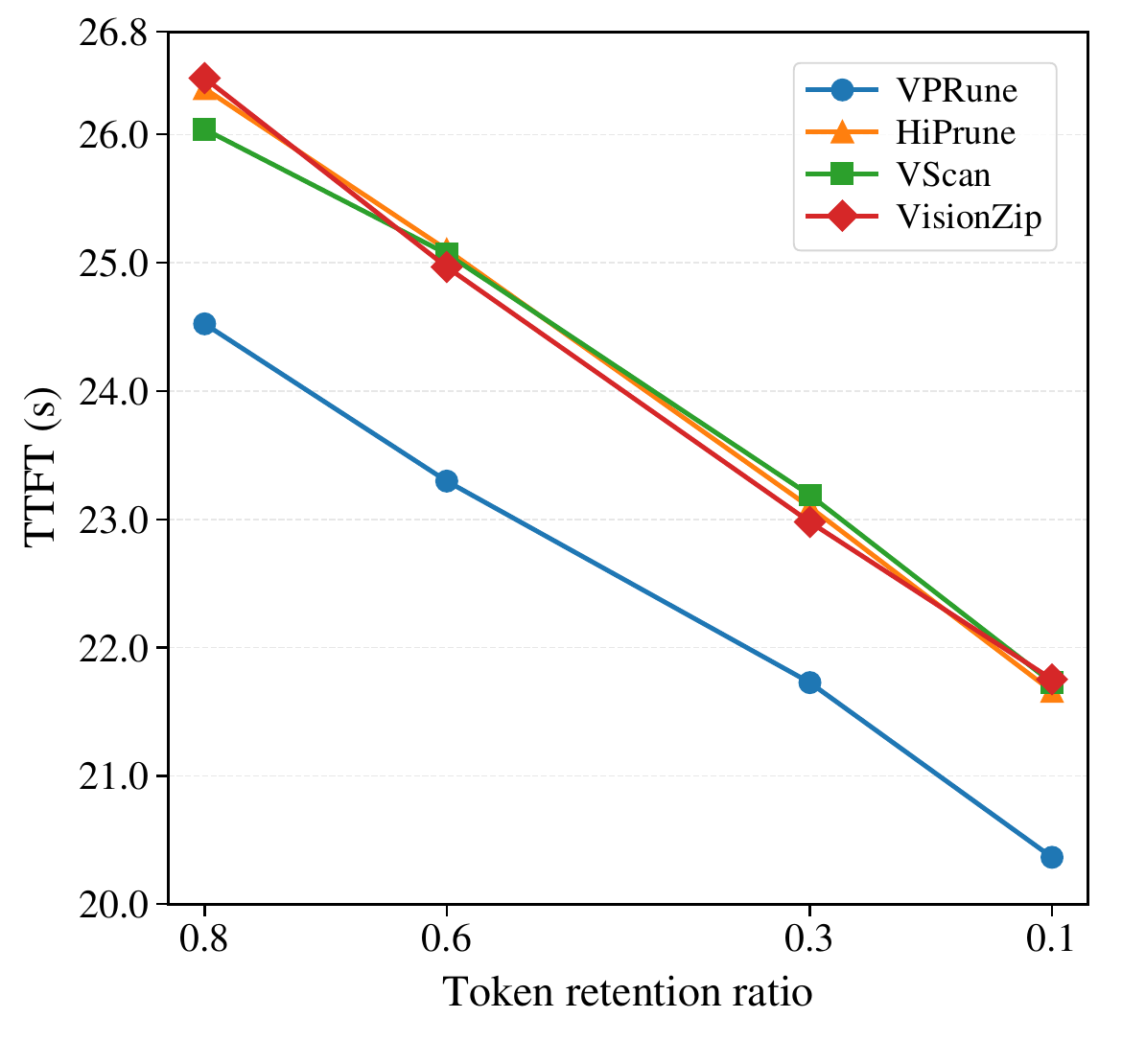}\\[-0.5mm]
        (b) End-to-end TTFT
    \end{minipage}

\caption{
Efficiency on the QCS8550 CPU-only edge platform. VPRune maintains
near-baseline vision-encoder latency in (a) and consistently reduces
end-to-end TTFT across token-retention ratios in (b).
}
    \label{fig:efficiency}
\end{figure}

We evaluate practical inference efficiency on the QCS8550 CPU-only
edge platform. As shown in Fig.~\ref{fig:efficiency}, VPRune performs
pruning after the multimodal projector and therefore maintains
near-baseline vision-encoder latency, while VScan, VisionZip, and
HiPrune incur higher vision-side overhead.

More importantly, reducing the visual sequence before LLM inference
provides consistent end-to-end benefits. VPRune reduces TTFT from
25.551\,s for the full-token baseline to 20.365\,s at $b=0.1$,
corresponding to a 20.30\% reduction. The reductions remain
14.97\%, 8.82\%, and 4.02\% at $b=0.3$, $0.6$, and $0.8$,
respectively. Its TTFT remains comparable to the fastest competing
methods while maintaining substantially higher accuracy under
aggressive compression, demonstrating a favorable
accuracy--latency trade-off for edge deployment.

\section{Conclusion}
\label{sec:conclusion}

In this work, we investigate the performance degradation of visual token
pruning under aggressive compression and propose \textbf{VPRune}, a
training-free pre-LLM framework for efficient LVLM inference. VPRune
addresses three key limitations of low-budget pruning through visual-only
diversity selection, similarity-guided token recycling, and
position-preserving restoration. Experiments on FastVLM-1.5B across five
vision-language benchmarks demonstrate that VPRune achieves a favorable
accuracy--compression trade-off, particularly under tight token budgets.
CPU-only edge evaluations further show reduced end-to-end latency
while maintaining strong task performance, supporting efficient
resource-constrained LVLM deployment.

\bibliographystyle{IEEEbib}
\bibliography{strings,refs}

\end{document}